\documentclass[sigconf]{acmart}
\usepackage{xcolor}
\usepackage{enumitem}
\usepackage{colortbl}
\usepackage{threeparttable}
\usepackage{makecell}
\usepackage{multirow}

\definecolor{promptred}{RGB}{220,38,38}
\definecolor{promptblue}{RGB}{32,108,194}

\AtBeginDocument{%
  }

\copyrightyear{2026}
\acmYear{2026}
\setcopyright{cc}
\setcctype{by}
\acmConference[KDD '26]{Proceedings of the 32nd ACM SIGKDD Conference on Knowledge Discovery and Data Mining V.1}{August 09--13, 2026}{Jeju Island, Republic of Korea}
\acmBooktitle{Proceedings of the 32nd ACM SIGKDD Conference on Knowledge Discovery and Data Mining V.1 (KDD '26), August 09--13, 2026, Jeju Island, Republic of Korea}
\acmPrice{}
\acmDOI{10.1145/3770854.3780294}
\acmISBN{979-8-4007-2258-5/2026/08}

\begin{document}

\title{HumanLLM: Towards Personalized Understanding and Simulation of Human Nature}

\author{Yuxuan Lei}
\orcid{0009-0006-3235-8674}
\affiliation{%
  \institution{University of Science and Technology of China}
  \streetaddress{Fuxing Road 100}
  \city{Hefei}
  \country{China}
  \postcode{230031}
}
\email{leiyuxuan@mail.ustc.edu.cn}

\author{Tianfu Wang}
\orcid{0000-0002-4386-3354}
\affiliation{%
  \institution{The Hong Kong University of Science and Technology (Guangzhou)}
  \streetaddress{No.1 Du Xue Rd}
  \city{Guangzhou}
  \country{China}
  \postcode{511455}
}
\email{twang566@connect.hkust-gz.edu.cn}

\author{Jianxun Lian}
\authornote{Corresponding author.}
\orcid{0000-0003-3108-5601}
\affiliation{%
  \institution{Microsoft Research Asia}
  \streetaddress{Danling Street 5}
  \city{Beijing}
  \country{China}
}
\email{jialia@microsoft.com}

\author{Zhengyu Hu}
\orcid{0009-0007-3097-9714}
\affiliation{%
  \institution{The Hong Kong University of Science and Technology (Guangzhou)}
  \streetaddress{No.1 Duxue Road, Nansha District}
  \city{Guangzhou}
  \country{China}
  \postcode{511453}
}
\email{zhu021@connect.hkust-gz.edu.cn}

\author{Defu Lian}
\orcid{0000-0002-3507-9607}
\affiliation{%
  \institution{University of Science and Technology of China}
  \streetaddress{Fuxing Road 100}
  \city{Hefei}
  \country{China}
  \postcode{230031}
}
\email{liandefu@ustc.edu.cn}

\author{Xing Xie}
\orcid{0009-0009-3257-3077}
\affiliation{%
  \institution{Microsoft Research Asia}
  \streetaddress{Danling Street 5}
  \city{Beijing}
  \country{China}
}
\email{xing.xie@microsoft.com}

\renewcommand{\shortauthors}{Yuxuan Lei et al.}

\begin{abstract}
Motivated by the remarkable progress of large language models (LLMs) in objective tasks like mathematics and coding, there is growing interest in their potential to simulate human behavior—a capability with profound implications for transforming social science research and customer-centric business insights. However, LLMs often lack a nuanced understanding of human cognition and behavior, limiting their effectiveness in social simulation and personalized applications. We posit that this limitation stems from a fundamental misalignment: standard LLM pretraining on vast, uncontextualized web data does not capture the continuous, situated context of an individual's decisions, thoughts, and behaviors over time. To bridge this gap, we introduce HumanLLM, a foundation model designed for personalized understanding and simulation of individuals. We first construct the Cognitive Genome Dataset, a large-scale corpus curated from real-world user data on platforms like Reddit, Twitter, Blogger, and Amazon. Through a rigorous, multi-stage pipeline involving data filtering, synthesis, and quality control, we automatically extract over 5.5 million user logs to distill rich profiles, behaviors, and thinking patterns. We then formulate diverse learning tasks and perform supervised fine-tuning to empower the model to predict a wide range of individualized human behaviors, thoughts, and experiences. Comprehensive evaluations demonstrate that HumanLLM achieves superior performance in predicting user actions and inner thoughts, more accurately mimics user writing styles and preferences, and generates more authentic user profiles compared to base models. Furthermore, HumanLLM shows significant gains on out-of-domain social intelligence benchmarks, indicating enhanced generalization. This work paves the way for more human-centric AI systems by advancing research in social simulation, developing personalized companions, enabling marketing intelligence through simulated customer feedback, and powering more realistic user simulation for recommender systems.
\end{abstract}




\maketitle

\section{Introduction}

In recent years, large language models (LLMs) have demonstrated remarkable proficiency across a wide range of professional benchmarks, often achieving or even surpassing human-level performance in areas such as code generation \cite{jiang2024survey}, math reasoning \cite{ahn2024large}, and text summarization \cite{zhang2025systematic}. 
Beyond these technical domains, LLMs have begun to exhibit intriguing, nascent human-like characteristics, including the ability to express personality traits \cite{jiang2023personallm,jiang2023evaluating} and a rudimentary understanding of theory of mind \cite{strachan2024testing, chen2024tombench}. This evolution has spurred significant interest in leveraging LLMs for applications that require simulating human behavior, such as intelligent non-player characters (NPCs), emotional companions, personalized assistants, and large-scale sociological simulations \cite{wan2024building, xu2024can,hewittpredicting,gurcan2024llm}. These advancements highlight the considerable potential of LLMs to capture and simulate facets of human nature, opening new frontiers in both AI and social science research.

 
However, a critical gap remains in the personalized and nuanced understanding of human cognition and behavior. Although LLMs excel in professional benchmarks, their training on vast, general-purpose web corpora~\cite{yang2025qwen3,dubey2024llama,abdin2024phi} is not optimized for simulating the intricacies of individual humans. This limitation stems from a fundamental misalignment: standard pretraining relies on disconnected text snippets, which fail to capture the continuous, situated context shaping an individual's decisions, thoughts, and behaviors over time. Consequently, when simulating human feedback or actions in complex social scenarios, these models often struggle to accurately predict motivations, infer inner states, or forecast future actions. This challenge motivates our central research question: Can we build a universal foundation model for deeper, more personalized understanding and simulation of human cognition and behavior?


Recent research has explored pathways to make LLMs more human-like. Some studies have focused on training models to predict outcomes from controlled psychology experiments or economic games \cite{xie2025fm, binz2025foundation, hewittpredicting}. Others have utilized role-playing benchmarks based on fictional characters \cite{wang2025coser}. While valuable, the data used in these approaches are often limited in scale, collected in artificial settings that emphasize group averages, or derived from fictional narratives. Consequently, they may not fully capture the complexity, spontaneity, and long-term dynamics of authentic human experiences as they unfold in the real world.

To address these limitations, we propose a novel approach: harnessing large-scale, real-world user data from diverse online platforms to ground the learning of human nature in authentic experiences. Platforms like Reddit, Twitter, Blogger, and Amazon host a wealth of unstructured user-generated content that reflect the genuine, spontaneous behaviors and expressed thoughts of millions of individuals over extended periods. This data provides an unprecedented resource for modeling the intricate relationship between a person's identity ($P$), their environment ($E$), and their resulting actions ($B$), as captured by Lewin's seminal equation, $B = f(P, E)$ \cite{lewin2013principles}. Nevertheless, the critical challenge lies in transforming these raw, noisy logs into a well-narrative form that chains an individual's scattered behaviors into a coherent trajectory for model learning. 
Inspired by this, we introduce HumanLLM, a foundation model tailored for the personalized understanding and simulation of human individuals. The core of our approach is the construction of the Cognitive Genome Dataset. We collect real-world user data from multiple online platforms and employ a rigorous, three-stage pipeline—comprising data filtering, data synthesis, and data quality control—to automatically curate over 5.5 million user logs. This process distills the raw data into high-quality representations of user profiles, situational scenarios, and associated social question-answer pairs, effectively capturing patterns of behavior and thought. Based on this dataset, we design a suite of diverse learning tasks, including profile generation, social question answering, and writing imitation, and apply supervised fine-tuning to equip the model with the ability to predict a wide range of individualized human behaviors and cognition. To mitigate catastrophic forgetting of general capabilities, we adopt a model merging strategy \cite{xiao2023lm}, blending the fine-tuned model with its original base, which proves more effective than joint fine-tuning with general instruction data.


Our comprehensive evaluations demonstrate that HumanLLM achieves superior performance on in-domain tasks, significantly outperforming base models in predicting user actions, inner thoughts, and stylistic imitation. Furthermore, HumanLLM shows enhanced generalization capabilities, delivering substantial gains on out-of-domain social intelligence benchmarks like MotiveBench \cite{yong2025motivebench}  and TomBench \cite{chen2024tombench} . Through real-world application studies in profile generation, human behavior explanation, and personalized writing, we confirm that HumanLLM serves as a superior social data generator, human explainer, and user simulator.

Our main contributions are as follows:
\begin{itemize}[leftmargin=1.5em]

    \item To the best of our knowledge, we are the first to leverage large-scale, real-world user data from multiple online platforms to enable LLMs to learn personalized human behaviors and thoughts. We construct the Cognitive Genome Dataset through a rigorous curation pipeline, resulting in a rich resource of hundreds of thousands of users, millions of scenarios, and social QA pairs.

    \item We design a series of training tasks to help LLMs learn a wide range of personalized user behaviors and thoughts, resulting in HumanLLM—a foundation model with enhanced social intelligence.

    \item We conduct comprehensive evaluations on in-domain test data, public social benchmarks, and real-world applications, demonstrating that our model is a superior social data generator, human explainer, and user simulator. Code is available at \url{https://aka.ms/humanllm}

\end{itemize} 

\begin{figure*}[t] 
 \setlength{\abovecaptionskip}{3pt}
    \centering
    \includegraphics[width=\textwidth]{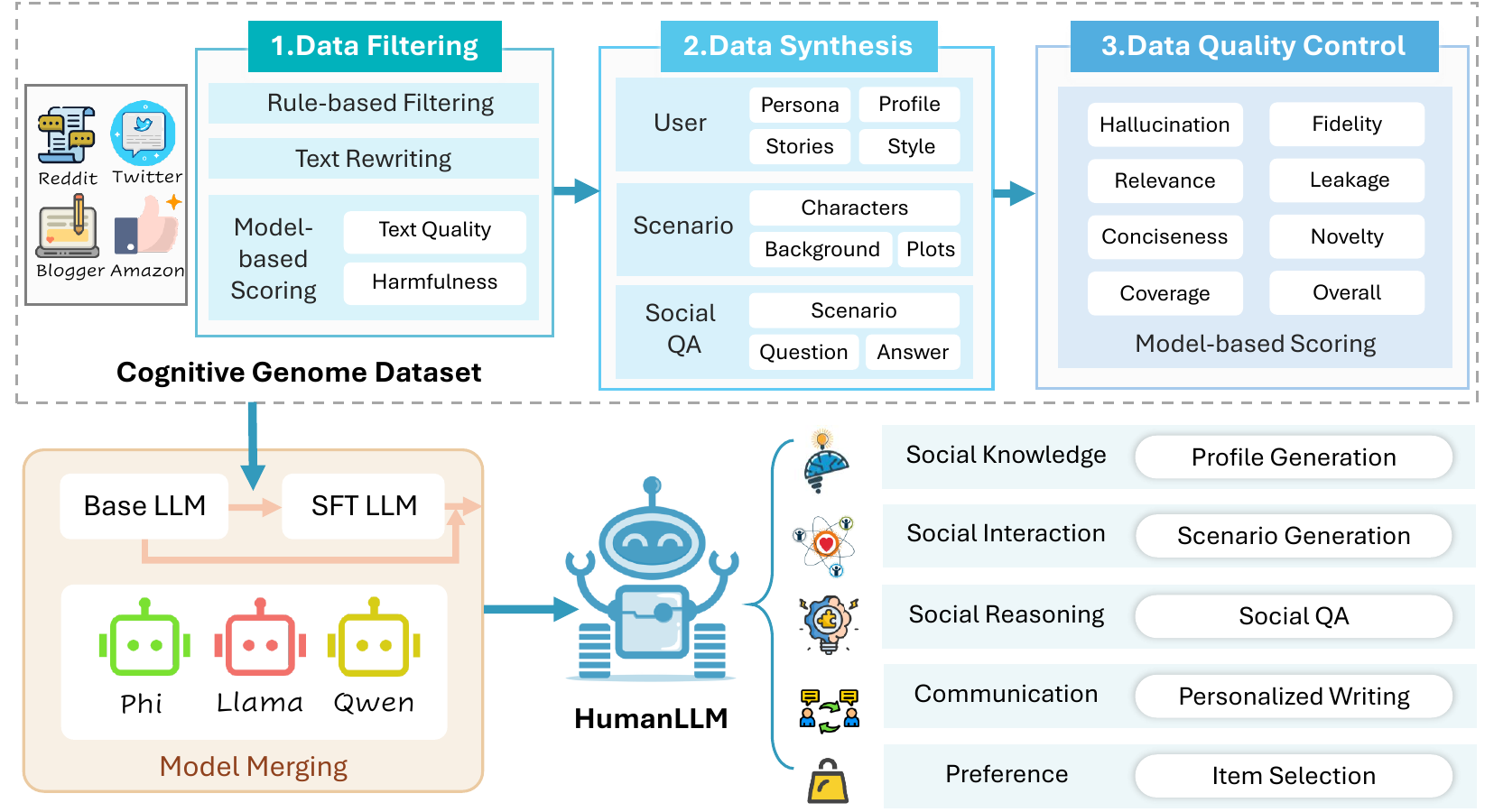}
    \vspace{-0.5cm}
    \caption{A graphical illustration for our pipeline.} 
    \label{fig:model}
\end{figure*}

\section{Cognitive Genome Dataset}
We select our data sources according to the following criteria: (1) each user must have multiple records; (2) the user population should be diverse; and (3) user records must reflect authentic, spontaneous behaviors and thoughts from everyday life. Based on these standards, we identify four high-quality sources: Reddit \cite{huggingfacegeclm-reddit-submissions} (47.2M submissions from 50 subreddits), Twitter \cite{enryu-twitter100m-tweets} (88.1M original tweets from CommonCrawl), Blogger \cite{schler2006effects} (681K blogs), and Amazon \cite{hou2024bridging} (571.54M item reviews). For details regarding the data sources, please refer to the Appendix \ref{appendix:data sources}. Based on these data sources, our data curation process consists of three stages: data filtering, data synthesis, and data quality control. We introduce each of these steps in detail in following sections.

\subsection{Data Filtering} The primary goal of this step is to clean the raw data, ensuring its authenticity, narrative richness, and safety.

We first apply rule-based filtering. Specifically, for Reddit data, we retain posts with $\text{score} \ge 2$, $\text{num\_comments}\ge 2$, and $\text{length} \ge 300$ characters, and filter out all posts from bots, moderators, deleted accounts, or system users. For Twitter data, we keep tweets with $\text{length} \ge 300$ characters, removing retweets, quote tweets, corporate or brand accounts, and advertisement-like content. For Blogger data, we exclude blogs shorter than 300 characters and those that do not contain first-person pronouns. For Amazon data, we filter out reviews shorter than 100 characters and apply a k-core filter, keeping only items with at least three purchase records and users with at least eight purchase histories in the dataset.

For the retained text, we use Llama-3.3-70B-Instruct for text rewriting, removing HTML tags, URLs, non-English content, and formatting noise. For Twitter, we additionally remove hashtags and mentions. Furthermore, we require the LLM to replace any sensitive or private information, such as home addresses or ID numbers, with fictitious data to protect user privacy.

Finally, we perform model-based scoring to further ensure quality and safety. Using Llama-3.3-70B-Instruct, each text is assigned two labels: a three-level quality label (high, medium, low) and a binary harmfulness label. High-quality content must contain detailed personal experiences, inner thoughts, and social behaviors, and avoid excessive advertising or noisy content. Harmful content includes anything involving violence, pornography, or privacy violations. We retain only high-quality and non-harmful text, resulting in a final dataset of 2.8 million Reddit posts, 673,000 tweets, 368,000 blogs, and 1.7 million reviews.

\subsection{Data Synthesis}
The core objective of this stage is to distill meaningful content from the raw data. Since the original user records inevitably contain some noise and are essentially unstructured textual logs, simply memorizing these texts does little to enhance a model’s social intelligence. Inspired by Lewin’s Equation \cite{lewin2013principles} from social psychology, $B = f(P, E)$, which posits that behavior ($B$) is determined jointly by the person ($P$) and their environment ($E$)—we recognize that understanding human behavior requires considering both individual characteristics and the situational context. Accordingly, we extract three types of content from the data: user, scenario, and social QA, which correspond to the components $P$, $E$, and $B$ in Lewin’s formulation. In this stage, we also employ Llama-3.3-70B-Instruct to synthesize the data. 

\textbf{User.} The user component focuses on providing a comprehensive, global, and long-term summary of an individual, encompassing their demographics, personality traits, core values, interests, emotional tone, and salient life experiences. This holistic profile serves as a foundation for understanding and predicting the user’s behaviors and thoughts. Specifically, for each user, we concatenate all available information (e.g., their entire sequence of Reddit posts) in chronological order and prompt the LLM to generate a corresponding profile. Importantly, because no one’s experiences can be fully captured and information gaps are inevitable, we choose to generate profiles in the form of natural paragraphs rather than structured templates. This approach avoids the common issues of missing values in structured profiles and preserves user-specific details that might otherwise be lost in rigid formats. We generate four distinct levels of user profiles to offer diverse perspectives and provide richer task types for LLM training:

\begin{itemize}[leftmargin=1.5em]

\item User Persona: A concise, short-form profile (less than 100 words) capturing a user’s essential identity and key personality traits.

\item User Profile: A more detailed and comprehensive profile (100–400 words) that includes background, interests, and summaries of significant experiences.

\item User Stories: A collection of the user’s important life experiences or events, with each story containing a summary and a detailed narrative describing a significant or unique occurrence. These stories reveal insights into the user’s cognition and behaviors.

\item Writing Style: A summary of the user’s distinctive writing style. Capturing linguistic patterns and stylistic tendencies enables the model to better imitate the user’s authentic voice and enhances personalized generation tasks.
\end{itemize}

\textbf{Scenario.} For every individual record, we aim to identify a complete story that includes the time, location, characters, and events, and organize this information into a structured triplet of background, characters, and plots. On one hand, these scenario data capture the full chain of user behaviors within specific contexts, enabling the LLM to learn to generate rich, contextually grounded content through scenario generation task. On the other hand, these structured scenarios also provide the necessary context for subsequent context-aware tasks within the social QA component.

\textbf{Social QA.} For the behavior ($B$) component, our goal is to enable the LLM to learn how individuals are likely to behave—both in terms of actions and inner thoughts—given their unique characteristics and the context of their situation. The most direct and effective way to achieve this is through question-answering tasks. Specifically, for each user record, we extract multiple scenario–question–answer triplets, with each question focusing on a different aspect: the user’s next action, inner mental state, or social reasoning. Unlike questions about next actions or inner mental states, social reasoning questions are designed to foster the LLM’s capacity for attribution—to explore why a user behaves in a certain way, rather than simply what they do or feel. Importantly, the scenario in these triplets is refined to ensure that the answer is not directly revealed, thus making the QA pairs more challenging and promoting deeper reasoning.


\subsection{Data Quality Control}
During the data synthesis process, we observe that despite strict prompt engineering, the LLM-generated content still exhibits certain issues. For example, some social QA pairs suffer from data leakage, where the answers could be directly copied from the scenario, making the pairs overly simplistic. Similarly, some generated profiles contain significant hallucinations, describing experiences that the user had not actually encountered. Therefore, a rigorous quality control process is essential. Given the scale of our data, we adopt an automated evaluation approach using Qwen-2.5-72B. Based on our analysis of the issues present in the synthesized data, we define several evaluation metrics, with each type of synthesized data requiring only a relevant subset of these metrics. Full metric description can be found in Appendix \ref{appendix:data quality control}.

\textbf{Hallucination.} All generated content must be faithful to the source text. This metric evaluates whether personas, profiles, stories, scenarios, and QA samples remain grounded in the main story and intent of the original records.

\textbf{Coverage.} This metric assesses how well the generated content—whether a persona, profile, story, scenario, or QA sample— captures the salient, meaningful, and unique aspects of the original user data.

\textbf{Conciseness.} This metric evaluates whether the generated persona or profile is succinct, cohesive, and free from redundancy or unnecessary repetition.

\textbf{Relevance.} This metric assesses whether the generated persona or profile focuses exclusively on user-relevant information, excluding generic, promotional, off-topic material from the original data.

\textbf{Fidelity.} This metric evaluates whether generated scenarios are vivid, detailed, and coherent rather than generic or superficial. It also assesses the logical connection and complementarity among scenario, question, and answer.

\textbf{Novelty.} This metric assesses how interesting, distinctive, and memorable the generated content is. It measures whether QA samples, profiles, stories, and scenarios highlight the user’s unique personality, experiences, and perspectives, evoking specific personal or social resonance rather than being formulaic or generic.

\textbf{Leakage.} This metric measures the extent to which the answer is directly revealed or obvious from the scenario description. 

\textbf{Overall.} This metric provides a holistic assessment of the overall quality of the generated content.

Each metric is scored on a scale from 1 to 10. We retain samples with an overall score greater than 8 and all other scores above 7. This step filters out approximately 30\% of users, 60\% of scenarios, and 40\% of social QA samples from each source. Complete data statistics are provided in Appendix \ref{appendix:statistics}. Overall, the synthesized data and the cleaned raw data together constitute our Cognitive Genome Dataset. The whole pipeline is illustrated in Figure \ref{fig:model}, which provides a comprehensive overview of each key component and the overall workflow.

\section{HumanLLM}
To train a powerful human-like foundation model, we first describe how we design training tasks based on the Cognitive Genome Dataset, and then present the specific implementation details for training HumanLLM.

\subsection{Task Design}
The Cognitive Genome Dataset encompasses a rich array of user behaviors and thoughts, providing a robust foundation for LLMs to learn the full spectrum of human nature. To enable the model to capture this diversity, we design six complementary learning tasks—profile generation, scenario generation, social question answering, writing imitation, personalized commenting, and item selection. Together, these tasks collectively span the major dimensions of human activity: self-representation, contextual interaction, social reasoning, communication, and preference-based decision making.

\textbf{Profile Generation.} Formally, given a brief user persona $p_a$, we prompt the LLM to generate a more detailed user profile $p_e$ as follows: $p_e = LLM(p_a)$. This task enables the model to infer a comprehensive user background and life trajectory from minimal information, thereby enhancing its social knowledge.

\textbf{Scenario Generation.} Formally, given a set of characters $c$ and background information $b$, we prompt the LLM to reconstruct the complete story plots: $plots=LLM(c, b)$. This task encourages the LLM to learn what constitutes a complete and authentic social scenario, as well as realistic social interactions.

\textbf{Social Question Answering.} Given a user persona $p_a$ or profile $p_e$, a scenario $s$, and a question $q$, we prompt the LLM to answer the question related to the user in that specific context: $a = LLM(p_a \,or\, p_e, s,q)$. This task enables the model to learn personalized user behaviors, thoughts, and motivations within specific scenarios, thereby strengthening its social reasoning abilities.

\textbf{Writing Imitation.} Given a set of text segments from the user's writing history $s_{hist}$, or a description of the user's writing style $w$, along with a new topic $t$, we prompt the LLM to generate a new text (e.g., post, blog, tweet, or review) in the user’s style for the given topic: $s_{new} = LLM(s_{hist}\,or\,w, t)$. This task improves the model’s capacity for personalized writing.

\textbf{Personalized Commenting.} Given a history of the user’s previous comments $c_{hist}$ and a new post $p_{new}$, we prompt the LLM to generate a comment that the user might write in response to the new post: $c_{new} = LLM(c_{hist}, p_{new})$. This task further enhances the model’s ability for personalized, context-aware writing.

\textbf{Item Selection.} Given $i_{hist}$, a sequence of items previously purchased by the user, and the user’s persona $p_a$ or profile $p_e$, we prompt the LLM to select the most preferred item from a set of candidates $i_{cand}$: $i_{target} = LLM(i_{hist}, p_a\,or\,p_e, i_{cand})$. This task allows the model to learn and simulate personalized user preferences.

Guided by Lewin’s Equation, our task design reflects a common cognitive chain in human behavior: recognizing the person, understanding the context, and generating contextually grounded actions, thoughts, or decisions. Each task captures a high-frequency instance of this process in real-world settings, enabling the model to learn how personal traits and situational factors jointly shape behavior. Moreover, the data are drawn from multiple platforms and cover diverse topics, ensuring both breadth and representativeness, and reinforcing the model’s ability to generalize across social contexts.

\subsection{Implementation Details}
Based on the Cognitive Genome Dataset, we construct training data covering six task types. Given the sheer volume of available data and the prohibitive computational cost of training, we sample each task to ensure roughly equal representation across tasks and data sources. This results in a final dataset comprising 1,195,717 training samples and 132,869 test samples. We organize the data in ShareGPT format and train our models using the next-token prediction task, masking out non-response positions from the loss calculation. To comprehensively demonstrate the effectiveness of our dataset, we train several mainstream open-source LLMs, including Qwen2.5-3B-Instruct \cite{team2024qwen2}, Qwen2.5-7B-Instruct \cite{team2024qwen2}, Qwen3-8B \cite{yang2025qwen3}, Phi-3-mini-128k-instruct \cite{abdin2024phi-}, and Llama-3.1-8B-Instruct \cite{dubey2024llama}. We use Llama-Factory \cite{zheng2024llamafactory} to implement model training. All models are fully fine-tuned on 8×40GB A100 GPUs, using a learning rate of 5e-6, a total batch size of 64, max sequence length of 8192, cosine learning rate scheduler, and a warm-up ratio of 0.5. We enable different stages of DeepSpeed Zero \cite{rajbhandari2020zero} based on model size. Training is conducted for 3 epochs, with the 8B models requiring approximately 120 hours. The Inference cost of HumanLLM  is identical to that of the base model, and we use the vLLM framework \cite{kwon2023efficient} for LLM inference, with the temperature set to 0.7.

It is important to note that catastrophic forgetting is a well-known issue in supervised fine-tuning (SFT) of LLMs, where the model tends to overfit to the new data at the expense of its general intelligence and generalization ability. A common remedy is to include a portion of general-purpose instruction tuning data during training. However, in our experiments, this approach yields suboptimal results: although models perform relatively well on the test set, their generalization to public social intelligence benchmarks is significantly degraded, and human evaluation of generations reveals a notable decline in response quality. Inspired by Lm-cocktail \cite{xiao2023lm}, we adopt an alternative approach—model merging—to mitigate catastrophic forgetting. Since the instruction-tuned version of base models already possesses strong general intelligence, we perform a 1:1 weight merge between the fine-tuned and original models, producing our final HumanLLM, which combines enhanced social intelligence with robust general capabilities. In our experiments, we report results using the merged models by default, and provide further analysis in the ablation study section.

\begin{table*}[t]
    \setlength{\abovecaptionskip}{0.5pt}  
    \caption{In-domain evaluation (\%) on core tasks. Models with HumanLLM augmentation show significant improvements. The "Imp." column reports the performance gains of HumanLLM over its corresponding base model. The best value and the second best for each column are marked in \textcolor{promptred}{red} and \textcolor{promptblue}{blue}, respectively.}
    \label{tab:in-domain}
    \centering
  \fontsize{9.3pt}{9.3pt}\selectfont
    \begin{tabular}{l|c c c c c|c|c}
        \toprule
        Model & 
        \makecell{Item\\Selection} & 
        \makecell{Profile\\Generation} & 
        \makecell{Scenario\\Generation} & 
        \makecell{Social\\QA} & 
        \makecell{Writing\\Imitation} & 
        Avg. & Imp. (\%) \\
        \midrule
        $\text{Centaur}$ & 4.36 & 3.58 & 5.40 & 0.17 & 1.14 & 2.93 & -- \\
        $\text{BE.FM}$ & 11.57 & 11.83 & 14.20 & 9.50 & 10.39 & 11.49 & -- \\
        $\text{GPT-4o}$ & 25.54 & 5.75 & 7.49 & 5.83 & 7.28 & 10.38 & -- \\
        \midrule
        Phi-3-mini-128k-instruct & 10.20 & 10.89 & 12.67 & 8.00 & 5.12 & 9.38 &  \\
        $\text{HumanLLM}_\text{phi-3}$ & 30.86 & 11.11 & 15.33 & 12.33 & 11.14 & 16.15 & 72.17 \\ \midrule
        Llama-3.1-8B-Instruct & 16.86 & 3.33 & 6.67 & 1.67 & 4.01 & 6.51 &  \\
        $\text{HumanLLM}_\text{Llama}$ & \textcolor{promptred}{36.56} & 15.33 & 16.33 & \textcolor{promptblue}{14.67} & \textcolor{promptblue}{26.06} & 21.79 & 234.72 \\ \midrule
        Qwen3-8B & 14.50 & 5.33 & 10.33 & 1.67 & 3.79 & 7.12 &  \\
        $\text{HumanLLM}_\text{Qwen3}$ & 32.20 & 31.78 & \textcolor{promptblue}{22.00} & 12.67 & 13.36 & 22.40 & 214.61 \\ \midrule
        Qwen2.5-3B-Instruct & 12.56 & 1.78 & 3.67 & 2.00 & 1.78 & 4.36 &  \\
        $\text{HumanLLM}_\text{Qwen2.5-3B}$ & 28.86 & 16.89 & 12.67 & 6.67 & 8.02 & 14.62 & 235.32 \\ \midrule
        Qwen2.5-7B-Instruct & 15.23 & 4.22 & 9.33 & 1.67 & 3.12 & 6.71 &  \\
        $\text{HumanLLM}_\text{Qwen2.5-7B}$ & 33.16 & \textcolor{promptred}{39.56} & \textcolor{promptred}{27.33} & 11.00 & 25.84 & \textcolor{promptred}{27.38} & \textcolor{promptred}{308.05} \\
        \midrule
        $\text{HumanLLM}^{\text{raw}}_{\text{Qwen2.5-7B}}$ & 19.44 & 8.67 & 10.67 & 1.33 & 8.02 & 9.63 & 43.52 \\
        $\text{HumanLLM}^{\text{gen}}_{\text{Qwen2.5-7B}}$ & \textcolor{promptblue}{35.73} & \textcolor{promptblue}{38.00} & 18.33 & \textcolor{promptred}{15.00} & \textcolor{promptred}{28.54} & \textcolor{promptblue}{27.12} & \textcolor{promptblue}{304.17} \\
        $\text{HumanLLM}^{\text{t1}}_{\text{Qwen3}}$ & 15.01 & 23.25 & 14.90 & 11.75 & 10.77 & 15.13 &112.50  \\
        $\text{HumanLLM}^{\text{t2}}_{\text{Qwen3}}$ & 31.97 & 4.58 & 8.36 & 6.58 & 13.58 & 13.01 & 82.72 \\
        
        \bottomrule
    \end{tabular}
\end{table*}

\section{Experiments}
\subsection{Overview}
Understanding and modeling human cognition and behaviors is a foundational challenge across the social sciences and artificial intelligence. Existing AI models are seldom optimized for social intelligence, which limits their effectiveness and adaptability in a wide range of social contexts. By training on the comprehensive and richly structured Cognitive Genome Dataset, our HumanLLM aims to transcend these limitations, enabling more robust, nuanced, and context-aware modeling of human cognition and behaviors.

To systematically evaluate the breadth and depth of our model’s social intelligence, we design a diverse suite of benchmark tasks that collectively reflect key dimensions of real-world human activity. Specifically, we first introduce an in-domain benchmark covering item selection, profile generation, scenario generation, social question answering, and writing imitation, spanning everything from individual preference modeling to social reasoning and expressive communication. To further assess generalization and the advancement of core social reasoning abilities, we conduct rigorous evaluations on two leading public social intelligence benchmarks. Finally, we illustrate HumanLLM’s practical utility and creative generative capacity through a series of real-world applications. Regarding experimental baselines, in addition to the direct comparison between HumanLLM and its base model, we introduce three additional models: Centaur~\cite{binz2025foundation}, BE.FM~\cite{xie2025fm}, and GPT-4o\footnote{The version we use is gpt-4o-2024-11-20.}\cite{hurst2024gpt}. The former two are also open foundation models for human behavior and cognition, both fine-tuned from the Llama-3.1-8B family, but they primarily focus on learning from human decision-making data such as psychological experiments, economic games, and survey responses. GPT-4o, in contrast, represents a state-of-the-art, high-capacity proprietary LLM.



\subsection{Results on In-domain Evaluation Tasks}
We evaluate HumanLLM and baseline models on five in-domain tasks: item selection, profile generation, scenario generation, social question answering, and writing imitation. To better quantify model performance, we convert open-ended generation tasks into multiple-choice questions with standard answers. Specifically, for the item selection task—which is naturally a multiple-choice problem—we require the model to choose its preferred item from 20 candidates. The ground truth is the actual item purchased by the user, while the remaining 19 negative examples are randomly sampled from the item pool. For the other four tasks, we formulate each as an eight-way multiple-choice question. That is, given a task instruction and eight candidate answers, we prompt the LLM to select the most appropriate response. The correct answer is the result extracted from real user data. For the seven negative options, we use responses generated by leading LLMs (GPT-4o \cite{hurst2024gpt}, Llama-3.1-8B-Instruct, Llama-3.1-70B-Instruct, Qwen2.5-7B-Instruct, Qwen2.5-72B-Instruct, Phi-4 \cite{abdin2024phi}, and Qwen3-8B) for each task. Since these models have not seen the user's actual behavior, their answers are considered suboptimal. Table~\ref{tab:in-domain} summarizes the results.
HumanLLM consistently outperforms its vanilla counterparts across all tasks and multiple base models. The best result is achieved by $\text{HumanLLM}_\text{Qwen2.5-7B}$ (27.38\% average). The largest improvements are observed on item selection and profile generation, suggesting that HumanLLM enhances the model’s capacity to understand user preferences and life trajectories. Improvements in writing imitation and scenario generation indicate better modeling of user style and context. These findings highlight the effectiveness of large-scale user-centric data and multi-task learning in improving social intelligence and personalization in foundation models. We also observe that the Centaur model, which is fine-tuned from the Llama-3.1-8B rather than its instruction-tuned version, tends to overfit to its training data and exhibits poor instruction-following capabilities. As a result, its performance falls short of expectations across all evaluated tasks. In contrast, BE.FM shows improved performance compared to Llama-3.1-8B-Instruct, but still lags significantly behind HumanLLM. Notably, in the in-domain evaluation setting, GPT-4o also underperforms HumanLLM, indicating that after targeted training, HumanLLM becomes more sensitive to real user behaviors and is better able to infer responses that align with user-specific characteristics.

\begin{table}[t]
    \setlength{\abovecaptionskip}{3.5pt}  
    \caption{OOD Performance on MotiveBench. The "Imp." column reports performance gains of HumanLLM over its base model. The best value and the second best for each column are marked in \textcolor{promptred}{red} and \textcolor{promptblue}{blue}.}
    \label{tab:motivebench}
    \centering
    \fontsize{9.3pt}{9.3pt}\selectfont
    \resizebox{0.5\textwidth}{!}{
    \begin{tabular}{l|ccc|c|c}
        \toprule
        Model & Amazon & Blogger & Persona & Avg. & Imp. (\%) \\
        \midrule
        $\text{Centaur}$  & 0.0366 & 0.0488 & 0.0266 &0.0373 & -- \\
        $\text{BE.FM}$  & 0.7300 & 0.7244 & 0.6061 & 0.6868 & -- \\
        $\text{GPT-4o}$ & 0.9011 & 0.7744 & 0.7744 & 0.8166 & -- \\
        \midrule
        Phi-3-mini-128k-instruct & 0.7244 & 0.6088 & 0.5933 & 0.6421 &  \\
        $\text{HumanLLM}_\text{phi-3}$ & 0.7422 & 0.6644 & 0.6311 & 0.6792 & 5.77 \\ \midrule
        Llama-3.1-8B-Instruct & 0.5866 & 0.6499 & 0.5216 & 0.5860 &  \\
        $\text{HumanLLM}_\text{Llama}$ & 0.7355 & \textcolor{promptred}{0.7388} & 0.6311 & 0.7018 & \textcolor{promptred}{19.76} \\ \midrule
        Qwen3-8B & \textcolor{promptblue}{0.8188} & 0.6655 & 0.6727 & 0.7190 &  \\
        $\text{HumanLLM}_\text{Qwen3}$ & \textcolor{promptred}{0.8222} & \textcolor{promptblue}{0.7033} & \textcolor{promptblue}{0.7027} & \textcolor{promptred}{0.7427} & 3.30 \\ \midrule
        Qwen2.5-3B-Instruct & 0.6511 & 0.5711 & 0.5800 & 0.5990 &  \\
        $\text{HumanLLM}_\text{Qwen2.5-3B}$ & 0.6744 & 0.6588 & 0.5833 & 0.6388 & \textcolor{promptblue}{6.64} \\ \midrule
        Qwen2.5-7B-Instruct & 0.7433 & 0.6366 & 0.6633 &0.6810 &  \\
        $\text{HumanLLM}_\text{Qwen2.5-7B}$ & 0.7677 & 0.6766 & 0.6866 & 0.7103 & 4.30 \\
        \midrule$\text{HumanLLM}^{\text{raw}}_{\text{Qwen2.5-7B}}$ & 0.7488 & 0.6711 & 0.6661 & 0.6953 & 2.10 \\
        $\text{HumanLLM}^{\text{gen}}_{\text{Qwen2.5-7B}}$ & 0.7544 & 0.6911 & 0.6688 & 0.7047 & 3.48 \\
        $\text{HumanLLM}^{\text{t1}}_{\text{Qwen3}}$ & 0.7988 & 0.6911& 0.6866 & 0.7254 & 0.89 \\
        $\text{HumanLLM}^{\text{t2}}_{\text{Qwen3}}$  &0.7966 & 0.6955 & \textcolor{promptred}{0.7138} & \textcolor{promptblue}{0.7353} & 2.26 \\
        \bottomrule
    \end{tabular}
    }
\end{table}

\begin{table*}[t]
    \setlength{\abovecaptionskip}{7.8pt}  
    \caption{OOD Performance on TomBench. NLC refers to Non-Literal Communication. The "Imp." column reports performance gains of HumanLLM over its base model. The best value and the second best for each column are marked in \textcolor{promptred}{red} and \textcolor{promptblue}{blue}.}
    \label{tab:tombench}
    \centering
    \fontsize{9.3pt}{9.3pt}\selectfont
    \begin{tabular}{l|cccccc|c|c}
        \toprule
        Model & Emotion & Desire & Intention & Knowledge & Belief & NLC & Avg. & Imp. (\%) \\
        \midrule
        $\text{Centaur}$ &0.4246&0.3686&0.4072 & 0.2820 & 0.3513 & 0.4164 & 0.3750 & -- \\
        $\text{BE.FM}$ &0.6514&0.5392&0.6658 & 0.4186 & 0.6190 & 0.6870 & 0.5969 & -- \\
        $\text{GPT-4o}$ &0.7524&0.6278&0.8235& 0.5779 & 0.8594 & 0.7861 & 0.7378 & -- \\
        \midrule
        Phi-3-mini-128k-instruct & 0.5585 & 0.5455 & 0.5245 & 0.4626 & 0.4662 & 0.5468 & 0.5173 & \\
        $\text{HumanLLM}_\text{phi-3}$ & 0.5725 & 0.5302 & 0.5706 & 0.4310 & 0.5347 & 0.6117 & 0.5418 & \textcolor{promptred}{4.74} \\ \midrule
        Llama-3.1-8B-Instruct & 0.5928 & 0.5448 & 0.6461 & \textcolor{promptred}{0.5240} & 0.6406 & 0.6526 & 0.6001 & \\
        $\text{HumanLLM}_\text{Llama}$ & 0.6135 & 0.5806 & \textcolor{promptblue}{0.6961} & 0.4664 & 0.6178 & 0.6608 & 0.6059 & 0.97 \\ \midrule
        Qwen3-8B & \textcolor{promptblue}{0.6468} & 0.6002 & 0.6639 & \textcolor{promptblue}{0.5039} & 0.6282 & 0.7922 & 0.6392 & \\
        $\text{HumanLLM}_\text{Qwen3}$ & \textcolor{promptred}{0.6614} & \textcolor{promptred}{0.6163} & \textcolor{promptred}{0.7230} & 0.4848 & 0.6624 & \textcolor{promptblue}{0.8054} & \textcolor{promptred}{0.6589} & 3.08 \\ \midrule
        Qwen2.5-3B-Instruct & 0.6089 & 0.5292 & 0.5211 & 0.4626 & 0.7758 & 0.7400 & 0.6063 & \\
        $\text{HumanLLM}_\text{Qwen2.5-3B}$ & 0.6289 & 0.5671 & 0.5721 & 0.3676 & 0.7468 & 0.7905 & 0.6122 & 0.97 \\ \midrule
        Qwen2.5-7B-Instruct & 0.6067 & 0.5864 & 0.5942 & 0.4198 & 0.7962 & 0.7135 & 0.6195 & \\
        $\text{HumanLLM}_\text{Qwen2.5-7B}$ & 0.6235 & 0.5655 & 0.6826 & 0.3873 & \textcolor{promptred}{0.8329} & 0.7568 & \textcolor{promptblue}{0.6414} & \textcolor{promptblue}{3.54} \\ \midrule
        $\text{HumanLLM}^{\text{raw}}_{\text{Qwen2.5-7B}}$ & 0.4888 & 0.4812 & 0.6000 & 0.4444 & \textcolor{promptblue}{0.8166} & 0.7000 & 0.5885 & -5.00 \\
        $\text{HumanLLM}^{\text{gen}}_{\text{Qwen2.5-7B}}$ & 0.4555 & 0.5250 & 0.6083 & 0.3777 & 0.5805 & 0.7000 & 0.5412 & -12.64 \\
        $\text{HumanLLM}^{\text{t1}}_{\text{Qwen3}}$ &0.6317&0.6027&0.6783&0.4619 & 0.6671&0.8043& 0.6410&0.28 \\
        $\text{HumanLLM}^{\text{t2}}_{\text{Qwen3}}$  &0.6278&\textcolor{promptblue}{0.6079}&0.6600&0.4781&0.6615&\textcolor{promptred}{0.8080}&0.6406&0.22 \\
        \bottomrule
    \end{tabular}
\end{table*}

\subsection{Results on Out-of-domain Benchmarks}
To evaluate the generalization of models, we evaluate HumanLLM and baseline models on MotiveBench \cite{yong2025motivebench} and TomBench \cite{chen2024tombench}, two out-of-domain social intelligence benchmarks. Both benchmarks provide a comprehensive assessment of social reasoning, encompassing a wide spectrum of human attributes such as motivation, emotion, desire, knowledge, belief, intention, and communication. Together, they reflect the essential facets of human cognition and interpersonal understanding that are fundamental for advanced AI systems. The results are provided in Table~\ref{tab:motivebench} and Table~\ref{tab:tombench}, respectively. HumanLLM outperforms its respective base models on all metrics. On MotiveBench, $\text{HumanLLM}_\text{Qwen3}$ achieves the highest average score (0.7427). On TomBench, the same model also leads (0.6589), particularly in intention and non-literal communication. Performance gains are consistent across different LLM backbones, indicating that the Cognitive Genome Dataset and our training approach improve LLM's social reasoning and generalization. These results suggest that HumanLLM can be broadly applied to social intelligence tasks beyond the original training distribution. In contrast, BE.FM exhibits unstable performance. It outperforms Llama-3.1-8B-Instruct on MotiveBench, but performs worse on TomBench, suggesting that training on data sources such as economic games and survey responses may provide only a partial view of human behavior and cognition, limiting the model’s ability to generalize across diverse behavioral settings. Given its substantially larger model capacity, GPT-4o outperforming HumanLLM on OOD benchmarks is expected.

\subsection{Ablation Study}
We aim to investigate three key questions: \textbf{Q1:} Does the carefully curated Cognitive Genome Dataset genuinely improve model performance? \textbf{Q2:} What are the advantages of the model merging approach compared to simply incorporating general-purpose instruction tuning data? \textbf{Q3:} Do all training tasks contribute to the model’s performance?

To address Q1, we directly train Qwen2.5-7B-Instruct on the original raw corpora (posts, tweets, blogs, and reviews) using a text completion task. Given the first half of a text, the model is tasked with generating the remaining half, with the split point randomly determined. The results of the trained model, noted as $\text{HumanLLM}^{\text{raw}}_{\text{Qwen2.5-7B}}$, are shown in Tables ~\ref{tab:in-domain}, ~\ref{tab:motivebench}, and ~\ref{tab:tombench}. We observe that training on the raw data yields only marginal improvements over the base model on in-domain and MotiveBench benchmarks, and actually results in a notable performance drop on TomBench. On one hand, the original data are fragmented and fail to capture users’ personalized behaviors and long-term trajectories. On the other hand, they contain substantial noise, making it difficult for models to learn meaningful behavioral patterns.

To answer Q2, we incorporate general-purpose instruction tuning data into the training. Specifically, we include the SmolTalk \cite{allal2025smollm2smolgoesbig} and Orca \cite{mitra2024agentinstruct} datasets, two widely used public SFT datasets covering more than twenty task types, including text editing, summarization, reasoning, mathematics, coding, long-context understanding, and more. We mix these datasets with our training data from the Cognitive Genome Dataset in a 0.25:0.25:0.5 ratio, aiming to preserve the model's generalization ability. The results of the trained model, noted as $\text{HumanLLM}^{\text{gen}}_{\text{Qwen2.5-7B}}$ shown in Tables ~\ref{tab:in-domain}, ~\ref{tab:motivebench}, and ~\ref{tab:tombench}, demonstrate that although the model's performance on the in-domain test set is comparable to that of the model merging approach, it falls significantly short on OOD benchmarks. Notably, on TomBench, it lags behind even the original base model by 0.0783. The reasons for this may be that model merging preserves the complementary strengths of both the base and SFT models by directly combining their weights, thus retaining both general and social intelligence and reducing catastrophic forgetting. Unlike joint fine-tuning on mixed data, merging reduces interference between different objectives and enables better out-of-domain generalization. What's more, the instruction-tuned base model has already undergone extensive, proprietary optimization by its original team using high-quality, undisclosed instruction data. In contrast, open-source instruction-tuning datasets may differ in domain and are often of lower quality, making them less effective at maintaining the base model's capabilities and preventing forgetting during joint fine-tuning.

To address Q3, and considering the training cost, we divide the six training tasks into two groups: Task Group 1 (Profile Generation, Scenario Generation, and Social Question Answering) and Task Group 2 (Writing Imitation, Personalized Commenting, and Item Selection). We then train Qwen3-8B using only the tasks within each group. The resulting models, denoted as $\text{HumanLLM}^{\text{t1}}_{\text{Qwen3}}$ and $\text{HumanLLM}^{\text{t2}}_{\text{Qwen3}}$, are reported in Tables~\ref{tab:in-domain}, \ref{tab:motivebench}, and \ref{tab:tombench}. The results indicate that, in in-domain evaluations, training on a specific task group leads to substantial performance improvements on the corresponding tasks, while performance on unseen tasks degrades compared to the full HumanLLM. Nevertheless, both models retain a certain degree of generalization ability and still slightly outperform the base Qwen3-8B. On OOD benchmarks, models trained on either Task Group 1 or Task Group 2 achieve performance levels between Qwen3-8B and HumanLLM, suggesting that both task groups contribute meaningfully and act synergistically in improving overall performance.

\begin{figure}[b]
    \centering
    \small    \includegraphics[width=1.0\columnwidth]{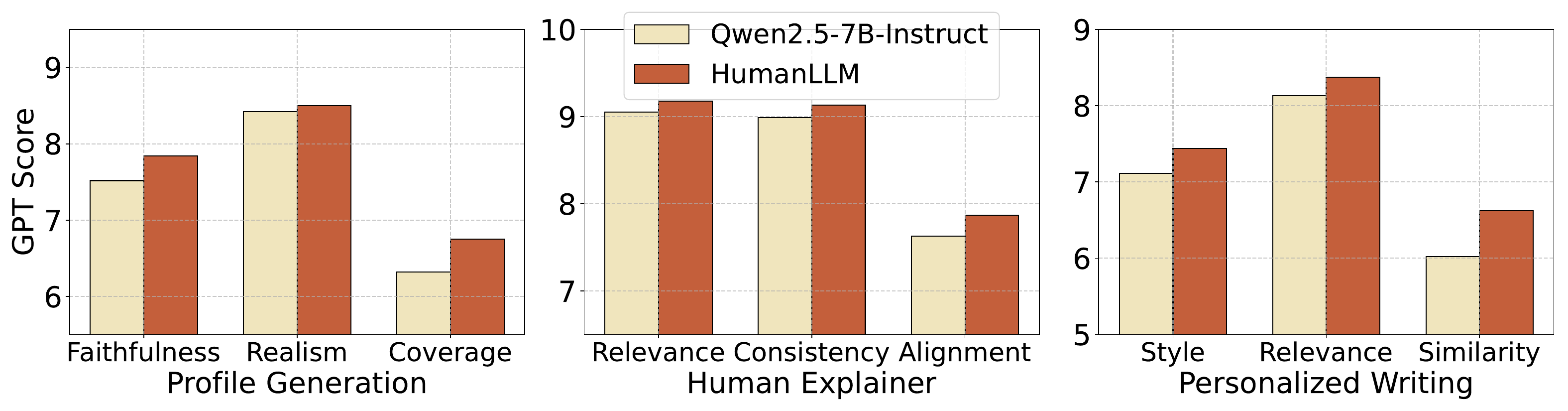} 
    \caption{Performance comparison between HumanLLM and its base model across three applications. Full results can be found in Appendix \ref{appendix:gpt results}.}
    \label{fig:applications_bar}
\end{figure}
\subsection{Real-World Application Studies}
We further explore the value of HumanLLM across three real-world applications: profile generation, human explainer, and personalized writing. We first conduct a systematic evaluation of HumanLLM’s generation quality using GPT-4o, with 1,600 samples per application. We then select several representative cases to provide an intuitive comparison of generation quality between models.

\textbf{Profile Generation.} Generating realistic and richly detailed user profiles is crucial for real-world applications such as data synthesis and user simulator. High-quality profiles enhance the diversity of synthetic data and authenticity of simulated user behaviors. To evaluate profile quality, we score generated profiles on faithfulness (how well the profile expands on the original persona), realism (realistic, coherent, and consistent with how a real person might be described), and coverage (matching the depth and breadth of human references). Results in Figure \ref{fig:applications_bar} demonstrate that our model outperforms the base model across all three metrics. Case analysis in Figure \ref{fig:case study persona} further shows that our model avoids hallucinated demographic details and produces nuanced, faithful expansions, leading to profiles that are more consistent with real user characteristics.

\begin{figure}[t]
    \centering
    \includegraphics[width=1.0\columnwidth]{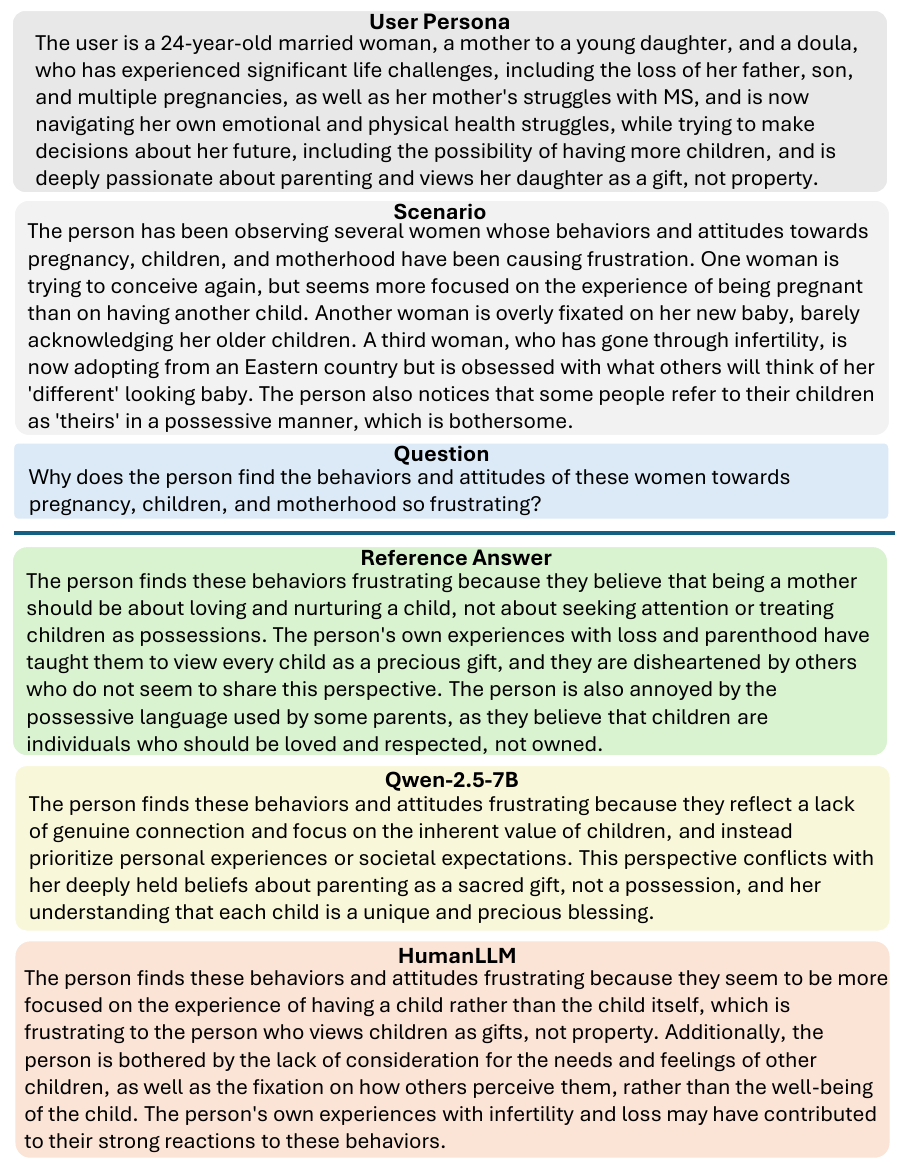} 
    \caption{Case Study for Human Explainer Application.}
    \label{fig:case study socail qa}
\end{figure}

\textbf{Human Explainer.} Understanding and explaining human behavior in specific contexts is crucial for advancing personal AI assistants, social simulation, and research on human cognition. This application shows models' ability to interpret users’ thoughts, motivations, and reactions to nuanced scenarios, allowing for more empathetic, context-aware, and personalized interactions. To evaluate the quality of generated explanations, we scores them on three key dimensions: contextual relevance (whether the answer is grounded in the user persona, scenario, and question), logical consistency (whether the reasoning is sound and coherent), and alignment with the reference (how well the answer matches the meaning and intent of real answer). Our results in Figure \ref{fig:applications_bar} show that HumanLLM consistently achieves high scores across these dimensions. For instance, in the provided case in Figure \ref{fig:case study socail qa}, HumanLLM’s answer stands out for its thorough reference to scenario details, its integration of the user’s background and perspective, and its comprehensive logical reasoning, leading to superior performance compared to the baseline model.

\textbf{Personalized Writing.} Personalized writing is a highly practical application in numerous real-world scenarios, such as personalized review writing,  automated email generation, and customer service communication. The ability to accurately imitate an individual user’s unique style, tone, and perspective not only enhances the authenticity and relevance of generated content but also improves user satisfaction and engagement. To evaluate the quality of generated writings, we score them on three metrics: style match (how well the generated text matches the user’s writing style, tone, and perspective), content relevance (the topical and logical coherence of the content), and content similarity (the extent to which the answer aligns with user generated text in terms of information coverage, depth, and main ideas). Results in Figure \ref{fig:applications_bar} show that HumanLLM delivers consistently better performance than the baseline across all metrics. Case analysis in Figure \ref{fig:case study writing} shows that HumanLLM’s output not only mirrors the user’s style but also faithfully reproduces critical details and reasoning patterns found in the reference review. This demonstrates the model’s strong ability to capture both stylistic nuances and substantive content in these applications.

\section{Related Work}

\subsection{The Need for Human-Centric LLMs}
LLMs such as Llama, Qwen, and Phi are pretrained on vast, general-purpose corpora, yet these sources rarely capture the intricacies of real human behaviors or the kinds of structured persona-scenario-behavior data necessary for advanced social simulation. This limitation has become increasingly relevant as LLMs are deployed in settings that require human-like understanding, such as AI companions, social simulation, and virtual agents~\cite{gurcan2024llm,wan2024building,xu2024can}. As LLMs are now widely adopted in applications that interact directly with humans, the importance of grounding their capabilities in authentic, diverse human experiences has grown more apparent. Recent works show that models excelling at academic or professional tasks often fail at social intelligence, underlining the distinction between academic and social capabilities in LLMs~\cite{xu2024academically}. These models may generate fluent and coherent text, yet still lack the nuanced reasoning needed to simulate individual personalities, motivations, or dynamic social contexts. Several studies have sought to bridge this gap: for example, using LLMs to predict outcomes of psychology or economics experiments~\cite{hewittpredicting}, or to serve as social participants in agent-based modeling~\cite{gurcan2024llm} and virtual environments~\cite{wan2024building}. Such efforts underscore the growing recognition that standard web-scale pretraining alone is insufficient for authentic social intelligence. Broader research trends, sometimes described as "machine psychology", increasingly advocate for new data collection methodologies, more sophisticated simulation tasks, and human-centered training regimes to align LLMs with the complexities of human cognition and behavior~\cite{hagendorff2023machine}.

\subsection{Measuring Human-Like Abilities in LLMs}
The need for systematic evaluation of LLMs’ human-like abilities, such as social intelligence, has given rise to numerous specialized benchmarks and datasets, as researchers seek to move beyond conventional metrics that focus mainly on linguistic or logical accuracy. Role-playing and theory-of-mind tasks are now commonly used to assess whether models can reason about others’ beliefs, intentions, and personalities~\cite{jiang2023personallm,jiang2023evaluating,strachan2024testing,chen2024tombench,wang2025coser}. Benchmarks like TOMBench~\cite{chen2024tombench}, Coser~\cite{wang2025coser}, SocialEval~\cite{shi2025socialeval}, and MotiveBench~\cite{yong2025motivebench} probe models’ abilities to maintain persona consistency, simulate established social roles, and perform human-like motivational reasoning. MotiveBench, in particular, targets the assessment of motivational reasoning, which is a key but often overlooked component of human-like intelligence. Meanwhile, works such as BigToM~\cite{wang2025simulate} and AgentSENSE~\cite{mou2025agentsense} examine LLMs’ social reasoning in both static and interactive, multi-agent settings, further expanding the scope of evaluation to more complex, real-world-like interactions. Several studies demonstrate that, despite rapid progress, even advanced LLMs lag significantly behind humans in theory-of-mind, emotional intelligence, and multi-party interaction~\cite{ying2025benchmarking,paech2024eq,li2024sid}. Comprehensive reviews and benchmarking efforts~\cite{ying2025benchmarking,li2024sid} emphasize that advancing LLMs’ social abilities requires not only better evaluation protocols but also richer training data that reflect the subtleties of everyday social cognition and interpersonal dynamics.

\subsection{Post-Training for Human Behavior Modeling}
To enhance LLMs’ modeling of individualized human behavior, recent research has explored targeted post-training approaches. Open foundation models such as BE.FM~\cite{xie2025fm} and Centaur~\cite{binz2025foundation} leverage large-scale behavioral and cognitive datasets to improve LLMs’ ability to predict, explain, and simulate human decisions. Personality conditioning through dialogue, as in Big5-Chat~\cite{li2024big5}, has been shown to help LLMs maintain consistent persona traits during interactions, resulting in more authentic and engaging conversational agents. SocialBench~\cite{chen2024socialbench} and related multi-party conversational datasets promote sociality and relational intelligence, while ToM-RL~\cite{lu2025tomrl} demonstrates that reinforcement learning from specialized curricula can significantly enhance LLMs’ theory-of-mind capabilities, even for relatively small models. Despite these promising advances, existing efforts often focus on optimizing LLMs for just some specified isolated social capabilities, such as personality expression or theory of mind, which can lead to overfitting and limited generalization. Broader and more integrated approaches remain rare. In contrast, our work aims to leverage large-scale post-training with diverse, real-world user data to comprehensively enhance LLMs’ social understanding, reasoning, and simulation abilities across a wide spectrum of human behaviors, ultimately moving toward the goal of universal, human-centric social intelligence in language models.

\section{Conclusion}

In this work, we introduce HumanLLM, a foundation model specifically designed to advance the personalized understanding and simulation of human cognition and behaviors. By curating the large-scale Cognitive Genome Dataset from multiple real-world online platforms, we enable the model to capture the nuanced dynamics of user behaviors and thoughts. Extensive experiments across in-domain tasks, out-of-domain benchmarks, and real-world applications demonstrate that HumanLLM is a superior social data generator, human explainer, and user simulator. These results highlight the potential of leveraging large-scale, user-centric data to drive meaningful progress in human-like AI. Looking forward, we believe that our approach opens promising directions for building AI systems that are not only more socially intelligent and empathetic, but also better equipped to understand, collaborate with, and serve real people across diverse scenarios.

\begin{acks}
The work was supported by grants from the National Natural Science Foundation of China (No. U24A20253).
\end{acks}

\bibliographystyle{ACM-Reference-Format}
\bibliography{myref}

\appendix

\renewcommand\thefigure{\Alph{section}\arabic{figure}} 
\setcounter{figure}{0} 
\renewcommand\thetable{\Alph{section}\arabic{table}} 
\setcounter{table}{0} 

\section{Data Details}
\begin{table*}[t]
  \centering
  \begin{threeparttable}
    \caption{Statistics of the Cognitive Genome Dataset at each data processing stage.}
    \label{tab:data-statistics}
      \begin{tabular}{l|r|r|rrr|rrr}
        \toprule
        \multirow{2}{*}{Source}
        & \multicolumn{1}{c|}{Raw} 
        & \multicolumn{1}{c|}{After DF}
        & \multicolumn{3}{c|}{After DS}
        & \multicolumn{3}{c}{After QC} \\
        \cmidrule(lr){2-2}
        \cmidrule(lr){3-3}
        \cmidrule(lr){4-6}
        \cmidrule(lr){7-9}
        & \multicolumn{1}{c|}{Texts}
        & \multicolumn{1}{c|}{Texts}
        & \multicolumn{1}{c}{Users}
        & \multicolumn{1}{c}{Scenarios}
        & \multicolumn{1}{c|}{Social QA}
        & \multicolumn{1}{c}{Users}
        & \multicolumn{1}{c}{Scenarios}
        & \multicolumn{1}{c}{Social QA} \\
        \midrule
        Reddit   & 47,200,000  & 2,800,000   & 102,861  & 612,012   & 898,264   & 75,661  & 259,470  & 549,383 \\
        Twitter  & 88,100,000  &   673,000   &  75,243  & 393,684   & 766,918   & 54,010  & 178,761  & 456,388 \\
        Blogger  &    681,000  &   368,000   &  41,607  & 1,048,883 & 1,012,066 & 30,738  & 448,020  & 271,210 \\
        Amazon   &571,540,000  & 1,700,000   & 210,000  &    --     &    --     &122,020  &    --    &    --   \\
        \midrule
        Total    &707,520,000  & 5,541,000   & 429,711  & 2,054,579 & 2,677,248 &282,429  & 886,251  &1,276,981 \\
        \bottomrule
      \end{tabular}
    \begin{tablenotes}
      \footnotesize
      \item \textit{Note:} ``DF'': Data Filtering, ``DS'': Data Synthesis, ``QC'': Quality Control.
    \end{tablenotes}
  \end{threeparttable}
\end{table*}
\subsection{Data Sources}\label{appendix:data sources}

\textbf{Reddit} \cite{huggingfacegeclm-reddit-submissions} (47.2M submissions from 50 subreddits): We use posts from active users on Reddit, a large-scale social platform organized by interest-based communities. Reddit posts capture a wide variety of personal opinions, experiences, and discussions.

\noindent\textbf{Twitter} \cite{enryu-twitter100m-tweets} (88.1M original tweets from CommonCrawl): We leverage tweets from a broad and diverse set of users on Twitter, a global microblogging platform. Tweets offer real-time, spontaneous expressions of users’ thoughts, emotions, and daily activities.

\noindent\textbf{Blogger} \cite{schler2006effects} (681K blogs): We utilize blog entries from the Blogger platform, where users share long-form personal stories and reflections. These blogs provide rich narratives of individual experiences, opinions, and life events.

\noindent \textbf{Amazon} \cite{hou2024bridging} (571.54M reviews): We extract user purchase histories from Amazon, including item names, ratings, and reviews. These records capture real-world consumer preferences and feedback, directly linked to personal decision-making and experiences.

\subsection{Data Statistics}\label{appendix:statistics}
Statistics of the Cognitive Genome Dataset at each data processing stage can be found in Table \ref{tab:data-statistics}.

\subsection{Data Quality Control}\label{appendix:data quality control}
To ensure the quality of the synthetic data, we design the following metrics for quality control.

\textbf{Hallucination.} All generated content must be faithful to the source text. Key details, events, and characterizations should be explicitly supported by the original user data or be reasonable, justifiable inferences. Minor details may be inferred if they do not alter the core narrative. This metric evaluates whether personas, profiles, stories, scenarios, and QA samples remain grounded in the main story and intent of the original records.

\textbf{Coverage.} This metric assesses how well the generated content—whether a persona, profile, story, scenario, or QA sample— captures the salient, meaningful, and unique aspects of the original user data. Good coverage requires that the most important and distinctive features from the source are distilled and included, with minimal omission of critical information.

\textbf{Conciseness.} This metric evaluates whether the generated persona or profile are succinct, cohesive, and free from redundancy or unnecessary repetition.

\textbf{Relevance.} This metric assesses whether the generated persona or profile focus exclusively on user-relevant information, excluding generic, promotional, off-topic material from the original data.

\textbf{Fidelity.} This metric evaluates whether generated scenarios are vivid, detailed, and coherent rather than generic or superficial. It also assesses the logical connection and complementarity among scenario, question, and answer, as well as the depth of contents and the preservation of emotional or personal resonance.

\textbf{Novelty.} This metric assesses how interesting, distinctive, and memorable the generated content is. It measures whether QA samples, profiles, stories, and scenarios highlight the user’s unique personality, experiences, and perspectives, evoking specific personal or social resonance rather than being formulaic or generic.

\textbf{Leakage.} This metric measures the extent to which the answer is directly revealed or obvious from the scenario description. Higher scores indicate that the answer requires genuine social reasoning or inference, while lower scores suggest that the answer can be simply copied or restated from the scenario with minimal reasoning.

\textbf{Overall.} This metric provides a holistic assessment of the overall quality of the generated content.

\subsection{Human Evaluation}\label{appendix:human eval}
During the data synthesis stage, the LLM is instructed to distill and summarize information from the original human-authored texts. In the subsequent data quality control stage, the synthesized data are further filtered using an LLM-as-judge framework to remove low-quality or inconsistent samples. To explicitly check whether the LLM introduces biases or hallucinations in the synthesized data, we conduct an additional round of human evaluation. Specifically, we stratify and sample 120 instances from the training set and ask three graduate students specializing in societal AI, none of whom are involved as co-authors, to assess the degree of hallucination and semantic plausibility. Each instance is independently evaluated by two annotators on a 1–5 Likert scale, where 5 indicates no hallucination and high semantic plausibility. The evaluation results show an average score of 4.65 and a Krippendorff’s $\alpha$ of 0.65, demonstrating that our synthesized data exhibit high reliability and strong fidelity to the original human behaviors.

\section{Application Results}\label{appendix:application}
\begin{table}[t]
    \setlength{\abovecaptionskip}{5.0pt}
    \caption{Performance on Profile Generation Application. The best value and the second best for each column are marked in \textcolor{promptred}{red} and \textcolor{promptblue}{blue}.}
    \label{tab:persona-generation}
    \centering
    \resizebox{0.99\columnwidth}{!}{
    \begin{tabular}{l|ccc|c|c}
        \toprule
        Model & 
        \makecell{Faith-\\Fulness} & 
        Realism & 
        Coverage & 
        Avg. & Imp. (\%) \\
        \midrule
        $\text{Centaur}$  & 5.7295 & 6.2459 & 4.4016 & 5.4590 & -- \\
        $\text{BE.FM}$  & 4.7500 & 6.1400 & 3.3508 & 4.7469 & -- \\
        $\text{GPT-4o}$ & 7.4166 & 8.8858 & 6.4483 & 7.5835 & -- \\
        \midrule
        Phi-3-mini-128k-instruct & 7.2333 & 8.5450 & 6.4858 & 7.4213 & \\
        HumanLLM$_{\text{phi-3}}$ & 7.7516 & 8.3833 & 6.6400 & 7.5916 & 2.30 \\ \midrule
        Llama-3.1-8B-Instruct & 6.6875 & 8.2333 & 5.9325 & 6.9511 & \\
        HumanLLM$_{\text{Llama}}$ & \textcolor{promptblue}{7.9500} & \textcolor{promptblue}{8.6166} & 6.8675 & 7.8113 & \textcolor{promptred}{12.37} \\ \midrule
        Qwen3-8B & 7.7616 & \textcolor{promptred}{8.8758} & \textcolor{promptblue}{6.9091} & \textcolor{promptblue}{7.8488} & \\
        HumanLLM$_{\text{Qwen3}}$ & \textcolor{promptred}{8.0258} & 8.7216 & \textcolor{promptred}{6.9166} & \textcolor{promptred}{7.8879} & 0.50 \\ \midrule
        Qwen2.5-3B-Instruct & 7.6966 & 8.4616 & 6.5066 & 7.5549 & \\
        HumanLLM$_{\text{Qwen2.5-3B}}$ & 7.7133 & 8.4016 & 6.5975 & 7.5707 & 0.21 \\ \midrule
        Qwen2.5-7B-Instruct & 7.5183 & 8.4250 & 6.3208 & 7.4213 & \\
        HumanLLM$_{\text{Qwen2.5-7B}}$ & 7.8391 & 8.4983 & 6.7500 & 7.6958 & \textcolor{promptblue}{3.70} \\
        \bottomrule
    \end{tabular}
    }
\end{table}

\begin{table}[t]
    \setlength{\abovecaptionskip}{5.0pt}
    \caption{Performance on Human Explainer Application. The best value and the second best for each column are marked in \textcolor{promptred}{red} and \textcolor{promptblue}{blue}.}
    \label{tab:human-explainer}
    \centering
    \resizebox{0.99\columnwidth}{!}{
    \begin{tabular}{l|ccc|c|c}
        \toprule
        Model & 
        \makecell{Contextual\\Relevance} & 
        \makecell{Logical\\Consistency} & 
        Alignment & 
        Avg. & Imp. (\%) \\
        \midrule
        $\text{Centaur}$  & 4.6200 & 5.9300 & 5.1010 & 5.2170 & -- \\
        $\text{BE.FM}$  & 8.3000 & 8.0375 & 6.6008 & 7.6461 & -- \\
        $\text{GPT-4o}$ & 9.5866 & 9.5675 & 8.3516 & 9.1685 & -- \\
        \midrule
        Phi-3-mini-128k-instruct & 8.9900 & 8.9575 & 7.5558 & 8.5011 & \\
        HumanLLM$_{\text{phi-3}}$ & 9.1125 & 9.0091 & 7.7708 & 8.6308 & 1.52 \\ \midrule
        Llama-3.1-8B-Instruct & 9.1633 & 9.1108 & 7.7833 & 8.6857 & \\
        HumanLLM$_{\text{Llama}}$ & 9.2291 & 9.1933 & \textcolor{promptblue}{7.9608} & \textcolor{promptblue}{8.7944} & 1.25 \\ \midrule
        Qwen3-8B & \textcolor{promptblue}{9.2541} & \textcolor{promptblue}{9.1983} & 7.8966 & 8.7830 & \\
        HumanLLM$_{\text{Qwen3}}$ & \textcolor{promptred}{9.2808} & \textcolor{promptred}{9.2241} & \textcolor{promptred}{7.9750} & \textcolor{promptred}{8.8266} & 0.50 \\ \midrule
        Qwen2.5-3B-Instruct & 8.9400 & 8.8041 & 7.4325 & 8.3922 & \\
        HumanLLM$_{\text{Qwen2.5-3B}}$ & 9.0375 & 8.9216 & 7.6283 & 8.5291 & \textcolor{promptblue}{1.63} \\ \midrule
        Qwen2.5-7B-Instruct & 9.0533 & 8.9900 & 7.6300 & 8.5577 & \\
        HumanLLM$_{\text{Qwen2.5-7B}}$ & 9.1775 & 9.1308 & 7.8675 & 8.7252 & \textcolor{promptred}{1.96} \\
        \bottomrule
    \end{tabular}
    }
\end{table}

\begin{table}[t]
    \setlength{\abovecaptionskip}{5.0pt}
    \caption{Performance on Personalized Writing Application. The best value and the second best for each column are marked in \textcolor{promptred}{red} and \textcolor{promptblue}{blue}.}
    \label{tab:personalized-writing}
    \centering
    \resizebox{0.99\columnwidth}{!}{
    \begin{tabular}{l|ccc|c|c}
        \toprule
        Model & 
        \makecell{Style\\Match} & 
        \makecell{Content\\Relevance} & 
        \makecell{Content\\Similarity} & 
        Avg. & Imp. (\%) \\
        \midrule
        $\text{Centaur}$  & 3.9270 & 4.9354 & 3.0453 & 3.9692 & -- \\
        $\text{BE.FM}$  & 6.4131 & 7.3355 & 5.0267 & 6.2584 & -- \\
        $\text{GPT-4o}$ & 8.6800 & 9.4091 & 7.6841 & 8.5910 & -- \\
        \midrule
        Phi-3-mini-128k-instruct & 6.6495 & 7.9285 & 5.6941 & 6.7573 & \\
        HumanLLM$_{\text{phi-3}}$ & 7.2321 & 8.2500 & 6.5066 & 7.3295 & \textcolor{promptred}{8.46} \\ \midrule
        Llama-3.1-8B-Instruct & 7.2792 & 8.3333 & 6.3246 & 7.3123 & \\
        HumanLLM$_{\text{Llama}}$ & \textcolor{promptred}{7.6822} & \textcolor{promptred}{8.6281} & \textcolor{promptred}{6.9701} & \textcolor{promptred}{7.7601} & \textcolor{promptblue}{6.12} \\ \midrule
        Qwen3-8B & 7.5306 & \textcolor{promptblue}{8.5766} & 6.6901 & 7.5991 & \\
        HumanLLM$_{\text{Qwen3}}$ & \textcolor{promptblue}{7.6257} & 8.5552 & \textcolor{promptblue}{6.9417} & \textcolor{promptblue}{7.7075} & 1.43 \\ \midrule
        Qwen2.5-3B-Instruct & 6.9751 & 8.0173 & 5.9205 & 6.9709 & \\
        HumanLLM$_{\text{Qwen2.5-3B}}$ & 7.2890 & 8.2444 & 6.4826 & 7.3386 & 5.28 \\ \midrule
        Qwen2.5-7B-Instruct & 7.1111 & 8.1333 & 6.0246 & 7.0896 & \\
        HumanLLM$_{\text{Qwen2.5-7B}}$ & 7.4382 & 8.3740 & 6.6259 & 7.4793 & 5.50 \\
        \bottomrule
    \end{tabular}
    }
\end{table}

\subsection{GPT-score Results}\label{appendix:gpt results}
Full results of the five backbone models across the three applications are presented in Tables \ref{tab:persona-generation}, \ref{tab:human-explainer}, and \ref{tab:personalized-writing}.

\subsection{Case Study}\label{appendix:case study}
\begin{figure}[h]
    \centering
    \includegraphics[width=1.0\columnwidth]{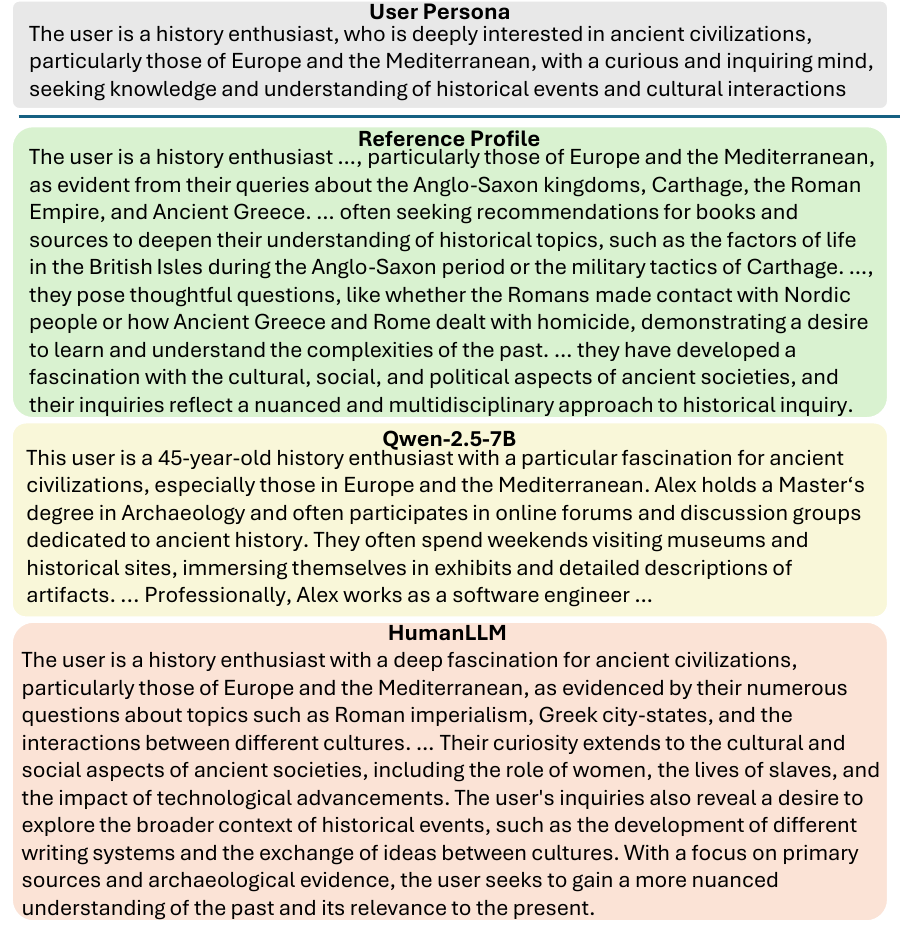} 
    \caption{Case Study for Profile Generation Application.}
    \label{fig:case study persona}
\end{figure}

\begin{figure}[h]
    \centering
    \includegraphics[width=1.0\columnwidth]{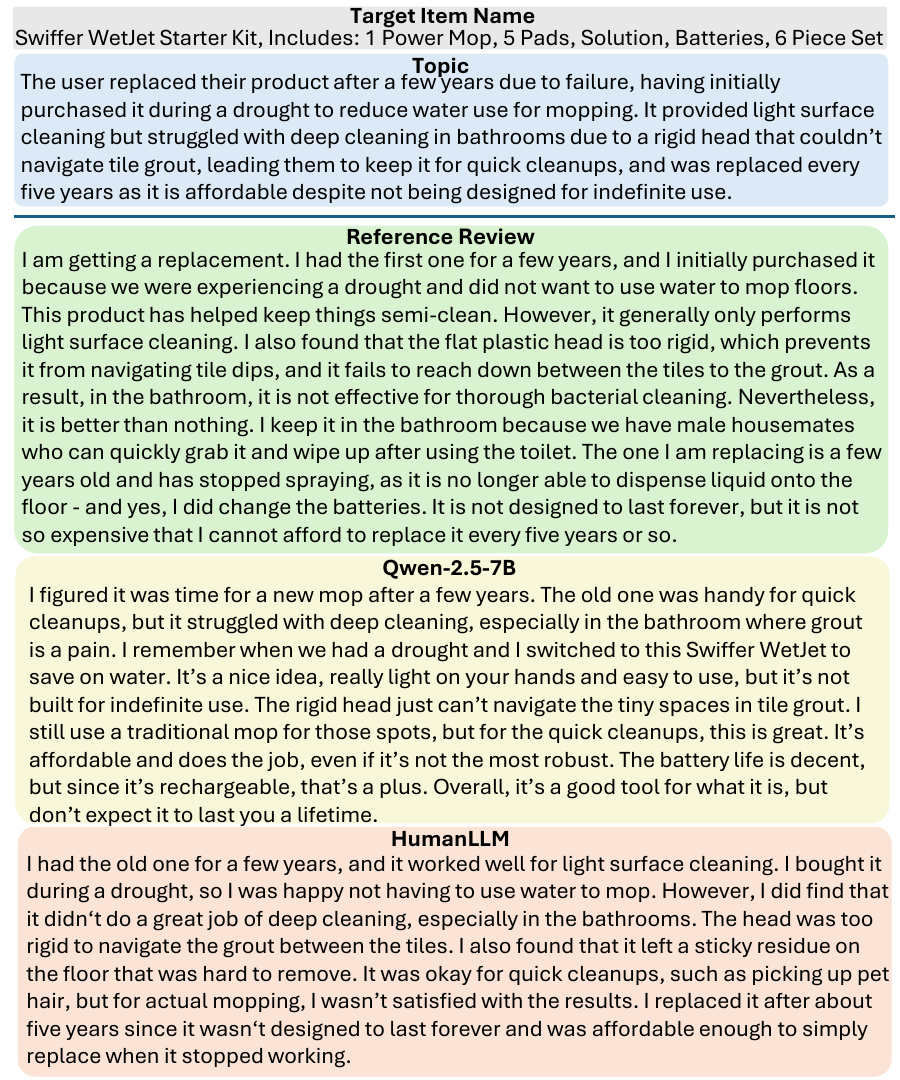} 
    \caption{Case Study for Personalized Writing Application.}
    \label{fig:case study writing}
\end{figure}

Due to space limitations in the main text, we present representative cases for the profile generation and personalized writing applications in this section, as shown in Figures \ref{fig:case study persona} and \ref{fig:case study writing}, respectively.

\onecolumn

\end{document}